\documentclass{article}
\usepackage{spconf,amsmath,graphicx}
\usepackage{xcolor}
\usepackage{color}
\usepackage{times}
\usepackage{epsfig}
\usepackage{graphicx}
\graphicspath{{images/}}
\usepackage{verbatim}
\usepackage{subfig}
\usepackage{array}
\usepackage{float}
\usepackage{multirow}
\usepackage{multicol}
\usepackage{siunitx}
\usepackage{hyperref}
\usepackage[english]{babel}
\usepackage[utf8]{inputenc}
\usepackage{algorithm}
\usepackage[noend]{algpseudocode}
\usepackage{booktabs}
\usepackage{colortbl}
\usepackage{enumitem,amssymb}
\usepackage{listings}
\usepackage{tablefootnote}

\usepackage[flushleft]{threeparttable}
\usepackage{sidecap,wasysym,array,tabularx,caption}
\usepackage{booktabs}
\definecolor{mygray}{gray}{.9}

\title{AV-Gaze: A Study on the Effectiveness of Audio Guided Visual Attention Estimation for Non-Profilic Faces}
\name{Shreya Ghosh$^{1}$ \qquad Abhinav Dhall$^{1,2}$ \qquad Munawar Hayat$^{1}$ \qquad Jarrod Knibbe$^{3}$}
  
\address{\small $^{1}$ \small{Monash University} $^{2}$ \small{Indian Institute of Technology Ropar} $^{3}$ \small{University of Melbourne}\\ 
\texttt{\small{\{shreya.ghosh, abhinav.dhall, munawar.hayat\}@monash.edu} jarrod.knibbe@unimelb.edu.au}}
\begin{document}
%
\maketitle
\begin{abstract}
In challenging real-life conditions such as extreme head-pose, occlusions, and low-resolution images where the visual information fails to estimate visual attention/gaze direction, audio signals could provide important and complementary information. In this paper, we explore if audio-guided coarse head-pose can further enhance visual attention estimation performance for non-prolific faces. Since it is difficult to annotate audio signals for estimating the head-pose of the speaker, we use off-the-shelf state-of-the-art models to facilitate cross-modal weak-supervision. During the training phase, the framework learns complementary information from synchronized audio-visual modality. Our model can utilize any of the available modalities i.e. audio, visual or audio-visual for task-specific inference. It is interesting to note that, when AV-Gaze is tested on benchmark datasets with these specific modalities, it achieves competitive results on multiple datasets, while being highly adaptive toward challenging scenarios.
\end{abstract}
\section{Introduction}
\label{sec:intro}
Estimating the human visual attention~\footnote{Here, we use the term gaze and visual attention interchangeably.\\ \textbf{Webpage:} \href{https://github.com/i-am-shreya/AV-Gaze}{https://github.com/i-am-shreya/AV-Gaze}} in terms of eye gaze is an important task in Image Processing, Computer Vision and Human Computer Interaction with applications in cognitive modelling, gesture recognition, assistive healthcare, human communication dynamics, and human-robot interaction~\cite{ghosh2021Automatic,ghosh2020automatic,joseph2020potential,borgestig2016eye,liao2020model}. Despite the major progress over the past few years, the RGB camera-based gaze estimation models fails in challenging scenarios such as occlusion, low resolution, and extreme head-pose. Previous works deal with these challenging cases by in-the-wild dataset based training~\cite{ghosh2022mtgls,zhang2020eth,kothari2021weakly}, unsupervised confidence mapping~\cite{chen2019unsupervised}, person-specific gaze redirection~\cite{yu2019unsupervised}, few-shot learning~\cite{park2019few} and temporal modelling~\cite{gaze360_2019}. Yet, due to the individuality of human face, such solutions could be error-prone especially when the face is fully/partially occluded or blurred. 

\begin{SCfigure}[][t]
\includegraphics[height=2.6cm, width=0.65\linewidth]{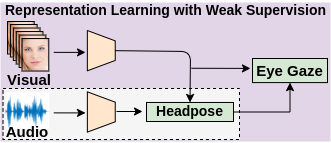}
\caption{\small We inv- estigate if audio guided head pose can improve visual attention estimation performance.}
\label{fig:eye_teaser}
\end{SCfigure}

\begin{figure*}[t]
    \centering
    \includegraphics[width=\linewidth]{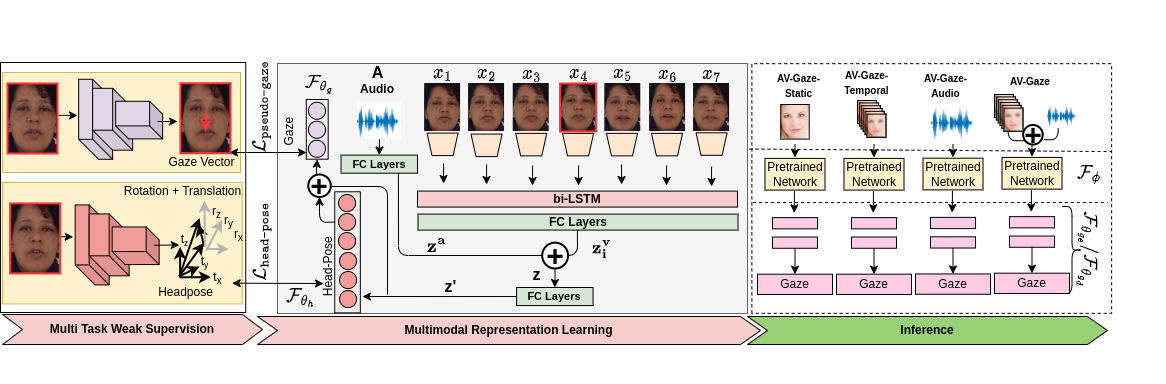}
    \vspace{-10mm}
    \caption{\small \textbf{An overview of our proposed AV-Gaze framework.} It learns gaze representations with weak supervision which is achieved by distilling knowledge from off-the-shelf deep models. The model is generalized well across different modality and tasks during inference.}
    \label{fig:pipeline}
    \vspace{-5mm}
\end{figure*}
\noindent On this front, we propose AV-Gaze, a different deep learning based solution that uses audio-visual signals for visual attention estimation. While the face and eye visibility play an important role in visual modality based visual attention/gaze and head-pose estimation, audio signals often captured concurrently with videos can be used for better head pose based guidance in challenging scenarios. Recent advancements in the speech community have also leveraged the aforementioned properties to detect head-pose and visual attention~\cite{greenwood2018joint,stefanov2019modeling}. However, these systems rely on the proper placement of audio sensor capturing devices and it mainly depend on handcrafted audio features~\cite{barnard2016audio} in different head-pose scenarios. In contrast to existing methods, we use joint learning from visual modality and transfer it to any uni-modal system to estimate the human visual attention in terms of gaze. Moreover, the past systems~\cite{greenwood2018joint,greenwood2017predicting} are tested on constrained datasets collected in a lab environment, therefore, the use of audio-visual modality to capture relevant information related to head-pose as well as achieving performance that is comparable to the State-Of-The-Art (SOTA) computer vision system has remained intractable. Fig.~\ref{fig:eye_teaser} shows an overview of the proposed framework. The problem formulation and multimodal training with weak supervision of our proposed AV-Gaze framework present different perspectives from vision-based multimodal eye gaze and head-pose estimation. In particular, it does not require time consuming and costly data collection and annotation. 
\textit{Overall, this work makes the following research contributions}: 
\begin{itemize}[topsep=1pt,itemsep=0pt,partopsep=1ex,parsep=1ex,leftmargin=*]

\item We propose \textbf{AV-Gaze}, a novel weakly supervised multimodal representation learning framework for gaze estimation by distilling knowledge from visual modality, guided by pseudo gaze and head-pose.

\item To the best of our knowledge, our model represents the first compilation of gaze/visual attention estimation using audio and video signals, which is especially impactful in challenging scenarios, especially where visual modality fails.

\item Through extensive experimental analysis, we show the impact of both audio and visual modalities in enabling automated head-pose guided gaze estimation.
\end{itemize}

\section{Multi-modal Representation Learning} \label{sec:method}


\noindent \textbf{Preliminaries.}
Lets assume AV-Gaze model $\mathcal{F}$ takes a video clip $\mathbf{I}$ as input which consists of video-frames and audio-clip $\mathbf{I} =$\{$\mathbf{X}, \mathbf{A}$\}, respectively (Refer Fig.~\ref{fig:pipeline}). $\mathcal{F}$ is parametrized by feature embedding module $\phi$ and label space module $\theta$.  In order to maintain the context information, we consider a frame chunk having 7 frames while the middle 4$^{th}$ frame is our point of concern. Lets denote these frames as $\mathbf{X} = \{x_1,x_2,\cdots x_i\}$, where $i = 7$. ResNet-18 framework based visual encoders $\mathcal{F}_{\phi-CNN}^{v}$ parameterized by $\phi-CNN$ maps the input frames into an intermediate representation $\mathbf{x_i^v} \in \mathbb{R}^{256}$ denoted as $\mathcal{F}_{\phi-CNN}^{v}:\mathbf{x_i} \to \mathbf{x_i^v}$. Consequently, it is passed through a bidirectional LSTM  $\mathcal{F}_{\phi-LSTM+FC}^{v}$ followed by FC layer as follows: $\mathcal{F}_{\phi-LSTM+FC}^{v}:\mathbf{x_i^v} \to \mathbf{z_i^v}$. Similarly, audio modality is also mapped to an intermediate $d$ dimensional representation $ \mathbf{z^a}\in \mathbb{R}^d$ via FC layers $\mathcal{F}_{\phi-FC}^{a}:\mathbf{A} \to \mathbf{z^a}$. Finally, the visual features $\mathbf{z_i^v}$ and audio features are concatenated to produce feature embedding $\mathbf{z}$. The embedding $\mathbf{z}$ is passed through the FC layers to produce latent feature $\mathbf{z'}$ which is then mapped to the cascaded output spaces $ \mathbf{h}\in \mathbb{R}^6 $ and $ \mathbf{g}\in \mathbb{R}^3 $, respectively by two sequential operations $\mathcal{F}_{\theta_h}:\mathbf{z'} \to \mathbf{h}$, and $\mathcal{F}_{\theta_g}:(\mathbf{h}\oplus \mathbf{z'})  \to \mathbf{g}$. Here, $\mathbf{h}$ and $\mathbf{g}$ are the head-pose and pseudo-gaze, respectively. The model $\mathcal{F}$ learns representation via joint learning of $(\mathcal{F}_\phi \circ \mathcal{F}_{\theta_h} \circ \mathcal{F}_{\theta_g})$.


\noindent \textbf{Audio Features.} We use an overlapping sliding frames over the audio signal in the time domain and compute log Filter Bank feature~\cite{greenwood2018joint}. The feature extraction parameters are as follows: the filter coefficient value, window size and overlap ratio are empirically chosen as 0.97, 10ms and 50\%, respectively. The resultant audio feature vector (40 filters) are calculated by applying triangular filters, on a Mel-scale to the power spectrum for extracting frequency bands. To maintain consistency, we normalise our input audio feature so that it has unit variance and zero mean.

\noindent \textbf{Multitask Weak Supervision.} 
\noindent \textbf{a) Head-Pose:}
As head-pose influence human visual attention, we use this for weak supervision in terms of 6-D vector extracted from `\texttt{img2pose}'~\cite{albiero2020img2pose} model. To train the network $(\mathcal{F}_{\phi} \circ \mathcal{F}_{\theta_h})$, the head-pose loss ($\mathcal{L}_{\texttt{head-pose}}$) is defined as $\mathcal{L}_{\texttt{head-pose}} = MSE(\mathbf{h},\mathbf{h'})$ where, $\mathbf{h}$ and $\mathbf{h'}$ are ground truth and predicted head-pose.

\noindent \textbf{b) Pseudo-Gaze:}
We use the eye-gaze estimator trained on ETH-Gaze dataset~\cite{zhang2020eth} to predict the pseudo labels of a head crop/facial image. Cosine similarity based gaze direction loss is used to train the $(\mathcal{F}_\phi  \circ \mathcal{F}_{\theta_h} \circ \mathcal{F}_{\theta_g})$ network. The loss ($\mathcal{L}_{\texttt{pseudo-gaze}}$) is defined as $\mathcal{L}_{\texttt{pseudo-gaze}} = \frac{\mathbf{g}}{||\mathbf{g}||_2}. \frac{\mathbf{g'}}{||\mathbf{g'}||_2},$ where, $\mathbf{g}$ and $\mathbf{g'}$ are ground truth and predicted gaze labels.

\noindent \textbf{AV-Gaze Training.} Given an input image ($\mathbf{x}$), the pseudo-gaze ($\mathbf{g}$) and head-pose ($\mathbf{h}$) labels are estimated via pre-trained models~\cite{albiero2020img2pose,zhang2020eth}. In absence of gaze information, we consider head-pose as the gaze/visual attention direction. These labels act as an ground truth label during representation learning phase. 
Our overall learning objective for AV-Gaze is as follows: $\mathcal{L}= \mathcal{L}_{\texttt{head-pose}} + \mathcal{L}_{\texttt{pseudo-gaze}} $

\noindent \textbf{Evaluation Protocol for AV-Gaze.} 
AV-Gaze takes synchronized facial/head-crop frames and audio chunks as input and learns visual attention/gaze representation via weak supervision. After representation learning, we use Linear Probing (LP) over the frozen backbone network for the relevant tasks for visual only, audio only and audio-visual modality (refer Fig.~\ref{fig:pipeline} \textit{Inference module}). More specifically for LP, the weights $\phi$ of the network is frozen and only $\mathcal{F}_{\theta}$ is updated. Note that we only adjust the task-specific output manifold (i.e. respective label spaces). 

\begin{SCtable}[][t]
\caption{\small Comparison on Gaze360 dataset in challenging settings in terms of mean angular errors~\cite{gaze360_2019}.}
\label{tab:challenging}
\resizebox{0.30\textwidth}{!}{
\begin{tabular}{l||c|c|c}
\toprule[0.4mm]
\rowcolor{mygray}
\multicolumn{1}{l||}{\textbf{Method}} & \multicolumn{1}{|c|}{\textbf{\begin{tabular}[c]{@{}c@{}}All\\ 360\textdegree\end{tabular}}}& \multicolumn{1}{|c|}{\textbf{\begin{tabular}[c]{@{}c@{}}Front\\ 180\textdegree\end{tabular}}} & \multicolumn{1}{|c}{\textbf{\begin{tabular}[c]{@{}c@{}}Front\\ Facing\end{tabular}}} \\ 
\hline \hline
Pinball-Static~\cite{gaze360_2019}* & 15.60 & 13.40 & 13.20 \\
AV-Gaze-Static & 15.55 & 13.10 & 13.23 \\ \midrule
Pinball-LSTM~\cite{gaze360_2019}* & 13.50 & 11.40 & 11.10 \\
AV-Gaze-Temporal & 13.40 & 11.17 & 11.00                             \\ \bottomrule[0.4mm]
\end{tabular}}
\vspace{-7mm}
\end{SCtable}

\section{Experiments and Results} \label{sec:exp_res}

\noindent \textbf{Experimental Settings.}
\noindent \textbf{1) Datasets:} We use approximately 50K samples (video chunks) extracted from AVSpeech dataset~\cite{ephrat2018looking}. During the frame extraction process, we choose the chunks in such a way that the standard deviation of the head-pose in the 7-frame chunk remains $\leq 0.1$ in the normalized label space. 
We further evaluate AV-Gaze on four benchmark datasets: \textit{CAVE}~\cite{smith2013gaze}, \textit{MPII}~\cite{zhang2017mpiigaze}, \textit{Gaze360}~\cite{gaze360_2019}, and \textit{DGW}~\cite{ghosh2021speak2label}. 
\noindent \textbf{2) Evaluation Settings:}
For a fair comparison, we follow the same evaluation protocols mentioned in the prior works for visual modality based gaze estimation methods~\cite{park2018deep,park2019few,yu2019improving,yu2019unsupervised,dubey2019unsupervised}. CAVE and MPII are image based datasets, Gaze360 contains temporal information, DGW contains audio-visual information. \textit{Thus, we apply AV-Gaze-Static for CAVE and MPII data; AV-Gaze-Static and AV-Gaze-Temporal for Gaze360 data and AV-Gaze-Static, AV-Gaze-Audio, AV-Gaze for DGW data.}  
\noindent \textbf{3) Baseline:} For visible and partially occluded faces, we compare AV-Gaze-Audio with OpenFace~\cite{baltrusaitis2018openface}, a SOTA vision-based library widely used for real-world tracking. Although it is a weak baseline as the gaze tracking method in OpenFace relies on landmark-detection performance which is challenging especially in the occluded scenario. 


\noindent \textbf{Quantitative Results.}
\noindent \textbf{1) Gaze Estimation for Non-profilic Faces:} As Gaze360 \cite{gaze360_2019} contains non-profilic faces, we validate the performance of AV-Gaze-Static and AV-Gaze-Temporal models for non-profilic faces on it (Refer \textbf{Table~\ref{tab:challenging}}\footnote{* denotes supervised methods in Table~\ref{tab:challenging},~\ref{tab:ssl_sota},~\ref{tab:audio_gaze} and~\ref{tab:down_dgw}.}). The angular errors correspond to the entire test set (All 360\textdegree) along with samples where the subjects are looking within 90\textdegree\ (Front 180\textdegree) and 20\textdegree\ (Front Facing). From Table~\ref{tab:challenging}, it is observed that the audio-guided head-pose followed by gaze estimation works well in non-frontal setting, i.e. Front 180\textdegree\ (within $\pm$ 90\textdegree). Quantitatively, it performs better than supervised baseline in both AV-Gaze-Static (13.40\textdegree\ \textrightarrow 13.10\textdegree, $\sim 2.23\%$) and AV-Gaze-Temporal settings (11.40\textdegree\ \textrightarrow 11.17\textdegree, $\sim 2.01\%$). The gain indicates that the knowledge distillation from audio feature guides head-pose which in turn improves the visual modality performance despite it is never trained in these visual scenarios. As the audio signal only provides the coarse head-pose even in absence of visual modality, thus, it rectifies gaze to a significant extend. 

\begin{SCtable}[][t]
\caption{\small\textbf{Visual on- ly Gaze Estimation.} Comparison of different gaze estimation methods on CAVE, MPII, and Gaze360 dataset in terms of Gaze Error (GE).}
\label{tab:ssl_sota}
\resizebox{0.30\textwidth}{!}{
\begin{tabular}{c||l|c}
\toprule[0.4mm]
\rowcolor{mygray} 
\multicolumn{1}{c||}{\textbf{Dataset}} & \textbf{Method}         & \multicolumn{1}{c}{\textbf{GE}} \\
\hline \hline
\multirow{5}{*}{CAVE}         & Jyoti et al.~\cite{jyoti2018automatic} *         & 2.22\textdegree                                                                                              \\  
                                       & Park et al.~\cite{park2018deep}*             & 3.59\textdegree                                                                                              \\  
                                       & Yu et al. \begin{tabular}[c]{@{}l@{}}~\cite{yu2019improving} \\   \end{tabular}& 3.54\textdegree                                                                                              \\  
                                       & Yu et al.~\cite{yu2019unsupervised}         & 3.42\textdegree                                                                                              \\  
                                       & AV-Gaze-Static                   & 3.55\textdegree                                                                                              \\ \midrule
\multirow{2}{*}{MPII}         & Park et al.~\cite{park2018deep}*             & 4.50\textdegree                                                                                               \\  
                                       & AV-Gaze-Static                    & 4.57\textdegree                                                                                               \\ \midrule
\multirow{2}{*}{Gaze360}      & Pinball-LSTM~\cite{gaze360_2019}*     & 13.50\textdegree                                                                                              \\ 
                                       & AV-Gaze-Temporal                    & 13.40\textdegree                                                                                             \\ \bottomrule[0.4mm]
\end{tabular}}
\vspace{-5mm}
\end{SCtable}

\begin{SCtable}[][b]
\caption{\small\textbf{Can Audio Signal Predict Gaze?} Comparison of audio gaze estimation with visual modality baselines. }
\label{tab:audio_gaze}
\centering
\resizebox{0.30\textwidth}{!}{
\begin{tabular}{c||l|c}
\toprule[0.4mm]
\rowcolor{mygray} 
\multicolumn{1}{c||}{\textbf{Dataset}} & \textbf{Method}         & \multicolumn{1}{|c}{\textbf{GE}} \\  
\hline \hline 
\multirow{2}{*}{AVSpeech}         &   AV-Gaze-Audio      &   7.6\textdegree                                                                                            \\  
                                       & OpenFace~\cite{baltrusaitis2018openface}           &      7.3\textdegree                                                                                       \\\midrule
\multirow{2}{*}{DGW}      &   AV-Gaze-Audio   &                                                                              60.44\%                 \\ 
                                       & Ghosh et al.~\cite{ghosh2021speak2label}*                   &                                         60.98\%                                                    \\ \bottomrule[0.4mm]
\end{tabular}}
\vspace{-7mm}
\end{SCtable}

\begin{figure*}[t]
    \centering
    \includegraphics[width=0.95\linewidth]{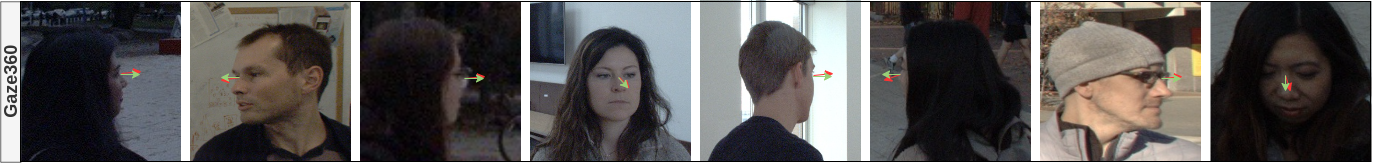}\\
    \includegraphics[width=0.95\linewidth]{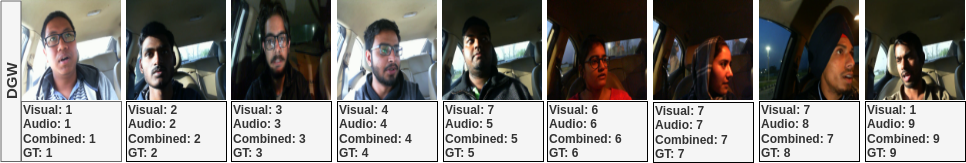}
    \vspace{-3mm}
    \caption{\small \textbf{Top:} Qualitative prediction results on Gaze360 dataset~\cite{gaze360_2019}. Here, red and green arrow represent predicted and ground-truth gaze direction, respectively. \textbf{Bottom:} Qualitative prediction results on DGW dataset~\cite{ghosh2021speak2label}. Here, the prediction of gaze zone in terms of 1-9 is reported in visual only, audio only and combined settings. GT stands for ground truth.}
    \label{fig:visual_qualitative}
    \vspace{-5mm}
\end{figure*}
\noindent \textbf{2) Visual-Only Gaze Estimation:} As the available datasets are estimating gaze from static visual modality, we evaluate the performance of the AV-Gaze-Static model for CAVE and MPII, and AV-Gaze-Temporal for Gaze360 dataset (Refer \textbf{Table~\ref{tab:ssl_sota}}). Although the main ambition for the AV-Gaze framework is to estimate visual attention/gaze in challenging scenarios, yet, it performs favourably for frontal face settings. 
The results indicate that AV-Gaze learns rich representation that is transferable across spatial and spatio-temporal settings.   

\noindent \textbf{3) Can Audio Signal Predict Gaze/Visual-Attention?}
To seek the answer to this question, we use the AV-Gaze-Audio model to estimate gaze/visual attention via audio signals with the corresponding gaze vector obtained using OpenFace~\cite{baltrusaitis2018openface} via visual modality (Refer \textbf{Table~\ref{tab:audio_gaze}}). For the AVSpeech dataset, we tested on random 5K samples disjoint from the training and validation samples. AV-Gaze performs competitively as compared to the OpenFace (7.6\textdegree\ \textrightarrow\ 7.3\textdegree). Similarly for the benchmark DGW dataset, the audio based gaze zone estimation accuracy of the audio-based AV-Gaze framework is 60.44\% as compared to the visual modality based prediction on the test set (i.e. 60.98\%). It is evident that given the audio modality, the gaze/visual attention estimation accuracy is almost as good as the vision-based predictions.

\noindent \textbf{4) Audio-Visual Gaze-Zone Estimation:}
\noindent We further evaluate the performance of AV-Gaze-Static and AV-Gaze framework on driver gaze zone prediction task~\cite{ghosh2021speak2label} as DGW dataset contains both audio and visual modality (Refer \textbf{Table~\ref{tab:audio_gaze}}). Our AV-Gaze-Static model outperforms the supervised baseline by $12.59\%$ and $13.12\%$ margin on validation and test set, respectively. Even with weighted k-NN on the frozen feature, our method outperforms the supervised baseline ($1.47\%$ on the validation set and $1.12\%$ on the test set). Note that a few methods~\cite{yu2020multi,lyu2020extract} outperforms our method as these utilize more task specific information such as body pose and other cues for driver gaze estimation as compared to our facial region based inference.

\begin{SCtable}[][t]
\caption{\small Performance com- parison with SOTA methods on driver gaze zone estimation on DGW dataset. Here, the LP refers to Linear Probing.}
\label{tab:down_dgw}
\centering
\resizebox{0.33\textwidth}{!}{
\begin{tabular}{l||c|c}
\toprule[0.4mm]
\rowcolor{mygray} 
\textbf{{Method}}& \textbf{\begin{tabular}[c]{@{}c@{}}Val\\ Acc. (\%)\end{tabular}} & \textbf{\begin{tabular}[c]{@{}c@{}}Test\\ Acc. (\%)\end{tabular}} \\
\hline \hline
Fridman et al.~\cite{fridman2015driver}*     &   {53.10}                                                                        &   {52.87}                                                                   \\ 
\begin{tabular}[c]{@{}c@{}}Vora et al.~\cite{vora2017generalizing}* \end{tabular} &  {56.25  }                                                                         &  {57.98 }                                                                   \\ 
SqueezeNet~\cite{iandola2016squeezenet}*                                                                             &  {59.53}                                                                           &  {59.18 }                                                                    \\ 
Baseline~\cite{ghosh2021speak2label}*                                                                              &  {60.10}                                                                           &  {60.98}                                                                    \\ 
Inception V3~\cite{DBLP:journals/corr/SzegedyVISW15}*                                                                           &  {67.93}                                                                           &  {68.04}                                                                    \\
Vora et al.~\cite{vora2018driver}*                                                         &  { 67.31}                                                       &  {68.12}                                                \\
ResNet-152~\cite{DBLP:journals/corr/HeZRS15}*                                                                              &  {68.94}                                                                           &  {69.01 }                                                                   \\
\begin{tabular}[c]{@{}c@{}}Yoon et al.~\cite{yoon2019driver}* \end{tabular}     &   {70.94}                                                                        &   {71.20}                                                                   \\

\begin{tabular}[c]{@{}c@{}}Stappen et al.~\cite{stappen2020x}*\\\end{tabular}     &   {71.03}                                                                        &   {71.28}                                                                   \\
\begin{tabular}[c]{@{}c@{}}Lyu et al.~\cite{lyu2020extract}*\\\end{tabular}     &   {85.40}                                                                        &   {81.51}                                                                   \\ 
\begin{tabular}[c]{@{}c@{}}Yu et al.~\cite{yu2020multi}* \\\end{tabular}     &   {80.29}                                                                        &   {82.52}                                                                   \\
\begin{tabular}[c]{@{}c@{}}AV-Gaze-Static (kNN)\end{tabular}                                                                   & 61.57                    & 62.10             \\
\textbf{AV-Gaze-Static (LP)}                                                                   & \textbf{72.69}                    & \textbf{74.10}             \\
\begin{tabular}[c]{@{}c@{}}AV-Gaze (kNN)\end{tabular}                                                                   & 67.49                    & 68.08             \\
\textbf{AV-Gaze (LP)}                                                                   & \textbf{75.40}                    & \textbf{76.48}             \\\bottomrule[0.4mm]
\end{tabular}}
\vspace{-7mm}
\end{SCtable}

\noindent \textbf{Ablation Studies.} 
\noindent \textbf{1) Impact of Pseudo Labels and Audio:}
To show the influence of different pseudo labels used in AV-Gaze framework, we conducted the ablation study where we train the AV-Gaze model with different possible combinations on Gaze360 and report the corresponding performance (Refer \textbf{Table~\ref{tab:ablation_auxiliary}}). From Table~\ref{tab:ablation_auxiliary}, it is observed that due to direct correlation the contribution of pseudo eye gaze (13.90\textdegree) is more compared to the head-pose (14.20\textdegree). Surprisingly, the audio modality performs competitively (15.70\textdegree) as compared to the visual modality. As expected, with the combination of the pseudo-gaze and head-pose the performance goes up to 13.73\textdegree and finally, the combination of all of these outperforms other (13.60\textdegree) by taking advantage of each module.
\noindent \textbf{2) How much temporal information does AV-Gaze need?}
Given a particular frame, we consider the contributions of both past and future contexts over the time duration. Our empirical estimation of optimal chunk size is 7. If we increase the chunk size (i.e. $>7$), the context information became irrelevant and noisy. We conduct a small experiment on random 5K samples of the AV-Gaze dataset where we progressively increase the context information and simultaneously predict the gaze. When the frame number is very less (i.e. $<7$) the corresponding audio signal is not significant. 

\noindent \textbf{Qualitative Results.}
\noindent \textbf{1) Visual Modality based Gaze Estimation:}
Fig.~\ref{fig:visual_qualitative} (Top) shows the qualitative analysis of our gaze prediction results using visual only modality. Similar to our initial hypothesis, AV-Gaze learns rich representation which can adapt in different challenging scenarios such as occlusion, image quality (i.e. low resolution) and extreme headpose. From the figure, it is also observed that even in absence of the eye-related information, the AV-Gaze can predict the gaze direction to some extent (refer Fig~\ref{fig:visual_qualitative} (Top) image 1$^{st}$, 3$^{rd}$, 5$^{th}$ and 6$^{th}$ from left). However, it is more aligned towards the head-pose of the person. 
\noindent \textbf{2) Audio-Visual Gaze Zone Estimation:}
Fig.~\ref{fig:visual_qualitative} (Bottom) demonstrate the prediction quality of our AV-Gaze framework using audio only, visual only and both audio-visual modalities. From the figure, it is observed that in most situations, the AV-Gaze can predict the gaze zone correctly. Please note that even in some cases, the audio channel is performing superior to the visual channel. Thus, it indicates that by leveraging multimodal signal, the visual attention of the concerned subject can be predicted.


\begin{SCtable}[][t] 
\caption{\small\textbf{Impact of Pseudo Gaze (PG), Head-Pose (HP) and Audio.} Results are presented in terms of Gaze Error (GE) (in \textdegree) on a subset of Gaze360 test dataset.}
\label{tab:ablation_auxiliary}
\centering
\resizebox{0.23\textwidth}{!}{
\begin{tabular}{c|c|c||c}
\toprule[0.4mm] 
\cellcolor{black!10} \textbf{PG} &\cellcolor{black!10} \textbf{HP} & \cellcolor{black!10} \textbf{Audio} & \cellcolor{black!10} \textbf{GE}\\
\hline \hline
$\checkmark$                 &                   &                        &         13.90\textdegree           \\\hline
                     & $\checkmark$              &                &  14.20\textdegree                \\ \hline
                     &                   & $\checkmark$             &              15.70\textdegree    \\ \hline
$\checkmark$                 & $\checkmark$              &                   &                 13.73\textdegree \\\hline
$\checkmark$                 & $\checkmark$              & $\checkmark$                      &         13.60\textdegree          \\
\bottomrule[0.4mm]
\end{tabular}}
\vspace{-7mm}
\end{SCtable}

\section{Conclusion and Future Work} \label{sec:conclusion}
\noindent In this work, we demonstrate a data-driven multimodal gaze estimation method that utilizes audio-visual modality to predict head-pose followed by gaze in a cascaded manner, especially where the visual modality fails. From both quantitative and qualitative analysis, we conclude that our proposed framework, AV-Gaze, learns a generic representation. Interestingly, the audio modality provides complementary information for predicting coarse head-pose/visual attention. However, the audio signal based solutions may have some limitations: \textit{First}, the position of the audio capturing device could effect the performance in real-time. \textit{Second}, several unavoidable environmental factors (i.e. background noise, interference of audio capturing devices) can affect the performance. Moreover, moving from single to multiparty interaction, the model performance will be highly dependent on proper speaker tracking and other challenges. Exploring more sophisticated models in multiparty interaction scenarios is left for future work. We believe that this work will commence an interesting research direction to explore audio modality as complementary information in challenging real-world settings for gaze/visual attention.

\bibliographystyle{unsrt}
\small \bibliography{AVGaze}

\end{document}